\documentclass[10pt,twocolumn,letterpaper]{article}

\usepackage[pagenumbers]{cvpr} %

\usepackage{lineno}

\newcommand*{\affaddr}[1]{#1} 
\newcommand*{\affmark}[1][*]{\textsuperscript{#1}}
\renewcommand{\thefootnote}{\fnsymbol{footnote}}

\usepackage{multirow} 
\usepackage{algorithmic,algorithm}
\usepackage{float}
\usepackage[accsupp]{axessibility}  %

\newcommand{\methodname}{ChronoDepth}
\newcommand{\method}{\texttt{\methodname}\xspace}

\newcommand\nonumfootnote[1]{%
\begingroup%
    \renewcommand\thefootnote{}\footnote{\hspace{-3.7pt}#1}%
    \addtocounter{footnote}{-1}%
\endgroup%
}

\usepackage{colortbl}

\definecolor{firstcolor}{rgb}{1, 0.6, 0.6}
\definecolor{secondcolor}{rgb}{1, 0.8, 0.6}
\definecolor{thirdcolor}{rgb}{1,1, 0.6}

\newcommand{\fst}[1]{\cellcolor{firstcolor}#1}
\newcommand{\snd}[1]{\cellcolor{secondcolor}#1}
\newcommand{\trd}[1]{\cellcolor{thirdcolor}#1}

\newcommand{\nE}{\mathbb{E}}

\definecolor{cvprblue}{rgb}{0.21,0.49,0.74}
\usepackage[pagebackref,breaklinks,colorlinks,allcolors=cvprblue]{hyperref}

\def\paperID{} %
\def\confName{CVPR}
\def\confYear{2025}

\title{Learning Temporally Consistent Video Depth from Video Diffusion Priors}

\author{
Jiahao Shao\affmark[1*] \quad
Yuanbo Yang\affmark[1*]\quad
Hongyu Zhou\affmark[1] \quad
Youmin Zhang\affmark[2, 6] \quad
Yujun Shen\affmark[4] \quad \\
Vitor Guizilini\affmark[5] \quad
Yue Wang\affmark[3] \quad
Matteo Poggi\affmark[2] \quad
Yiyi Liao\affmark[1$\dagger$] \vspace{2mm}\\
\affaddr{\affmark[1]Zhejiang University}  \quad  \affaddr{\affmark[2]University of Bologna} \quad \affaddr{\affmark[3]University of Southern California} \\
\affaddr{\affmark[4]Ant Group} \quad
\affaddr{\affmark[5]Toyota Research Institute} \quad
\affaddr{\affmark[6]Rock Universe AI}
}

\newcommand{\bz}{\mathbf{z}}

\newcommand{\bm}{\mathbf{m}}

\newcommand{\bd}{\mathbf{d}}

\newcommand{\cN}{\mathcal{N}}

\newcommand{\cR}{\mathcal{R}}

\newcommand{\cL}{\mathcal{L}}

\newcommand{\cE}{\mathcal{E}}

\makeatletter
\DeclareRobustCommand\onedot{\futurelet\@let@token\@onedot}
\def\@onedot{\ifx\@let@token.\else.\null\fi\xspace}

\makeatother

\renewcommand{\eqref}[1]{Eq.~\ref{#1}}
\newcommand{\tabref}[1]{Table~\ref{#1}}

\newif\ifcomment
\commenttrue
\ifcomment
	\newcommand{\ag}[1]{ \noindent {\color{red} {\bf Andreas:} {#1}} }
	\newcommand{\yl}[1]{ \noindent {\color{cyan} {\bf Yiyi:} {#1}} }
        \newcommand{\youmi}[1]{ \noindent {\color{gray} {\bf Youmin:} {#1}} }

\else
	\newcommand{\ag}[1]{}
	\newcommand{\yl}[1]{}
        \newcommand{\youmi}[1]{}
\fi

\begin{document}
\twocolumn[{%
		\renewcommand\twocolumn[1][]{#1}%
		\maketitle
            \vspace{-1.2cm}
		\begin{center}
			\includegraphics[width=0.90\textwidth,height=5.25cm]{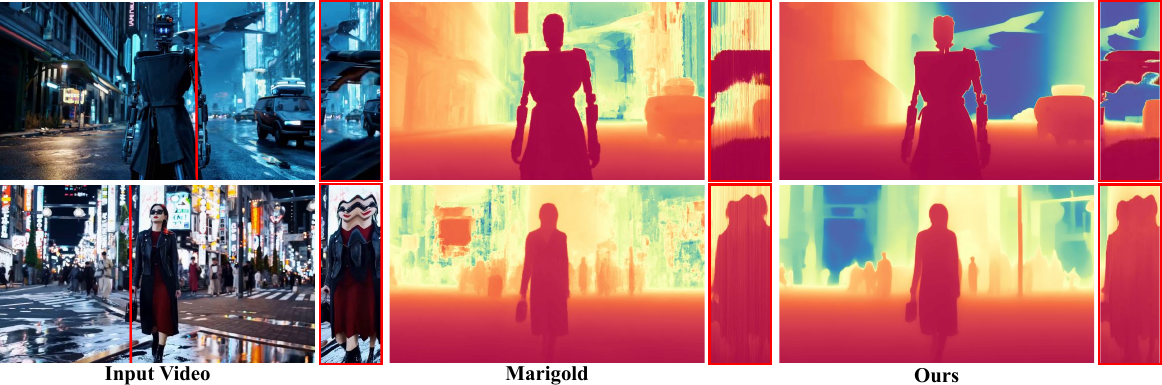}
\vspace{-0.3cm}			
   \captionsetup{type=figure}
			\captionof{figure}{
				\textbf{
					We present \method{}, a novel method derived from video diffusion model.} It can estimate video depth from arbitrary-length open-world videos while exhibiting high spatial accuracy and state-of-the-art temporal consistency.
			}\vspace{-0.1cm}
			\label{fig:teaser}
		\end{center}
	}]

\begin{abstract}
This work addresses the challenge of streamed video depth estimation, which expects not only per-frame accuracy but, more importantly, cross-frame consistency.
We argue that sharing contextual information between frames or clips is pivotal in fostering temporal consistency.
Therefore, we reformulate depth prediction into a conditional generation problem to provide contextual information within a clip and across clips.
Specifically, we propose a consistent context-aware training and inference strategy for arbitrarily long videos to provide cross-clip context. 
We sample independent noise levels for each frame within a clip during training while using a sliding window strategy and initializing overlapping frames with previously predicted frames without adding noise.
Moreover, we design an effective training strategy to provide context within a clip.
Extensive experimental results validate our design choices and demonstrate the superiority of our approach, dubbed \method{}.
    Project page:
    \href{https://xdimlab.github.io/ChronoDepth/}{xdimlab.github.io/ChronoDepth}.
    \nonumfootnote{$\ast$ denotes equal contribution. \hspace{1cm} $\dagger$ corresponding author.}%

\end{abstract}

\begin{figure*}[t!]
  \centering
  \includegraphics[width=0.9\linewidth]{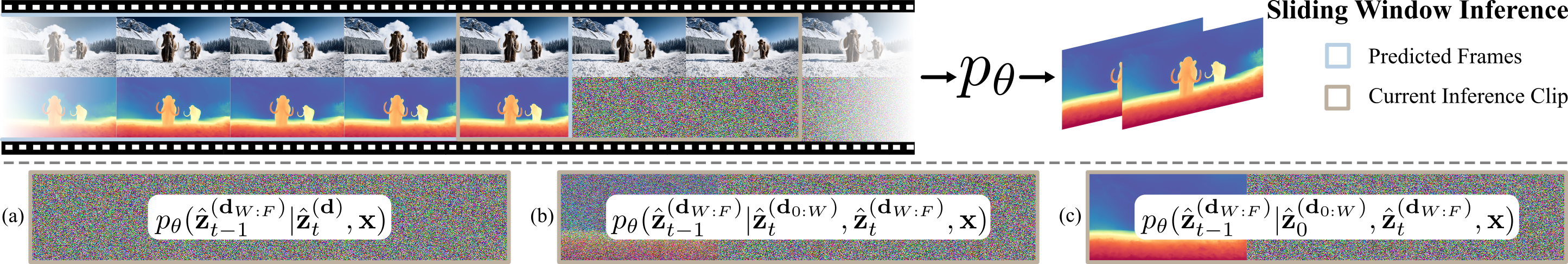}
  \vspace{-0.3cm}
  \caption{\textbf{
					Illustration of different inference strategies for infinitely long videos.} (a) Naive sliding window inference; (b) Inference with replacement trick; (c) Our proposed consistent context-aware inference. $F$, $W$ denotes number of clip and overlapping frames respectively.
                }
    \label{fig:infer_comp}
\vspace{-0.3cm}
\end{figure*}

\section{Introduction}
\label{intro}

Monocular image depth estimation methods have achieved significant advancements in recent years. Despite notable progresses in spatial precision~\citep{ranftl2020towards, yin2021learning, eftekhar2021omnidata, zhang2022hierarchical, ranftl2021vision, depthanything, Ke2023repurposing, fu2024geowizard, gui2024depthfm, xu2024diffusion}, these single-image depth estimation methods are built based on an i.i.d. assumption between frames. As there is no contextual information before or after each frame, this approach is inherently prone to flickering and temporal inconsistency. In the meantime, many applications in AR/VR, robotics, and video editing require an estimation of consistent depth over time, which calls for video depth estimation tools that exhibit both spatial accuracy and temporal consistency.

For arbitrarily long videos, given the purpose of autoregressive inference and the limitation of computing resources, it is necessary to split it into multiple smaller video clips to process them in a streamed manner.
Existing methods~\citep{wang2022less, 10233038} leverage the attention mechanism in a generalizable manner to aggregate contextual information for each video clip. However, such approaches cannot fully encode contextual information between clips. Therefore, flickering issues still persist across clips. 
Another line of methods leverages test-time training (TTT) paradigms~\citep{teed2018deepv2d, luo2020consistent, kopf2021robust, zhang2021consistent, wang2023neural} to alleviate this issue, wherein they fine-tune a single-image depth model on the testing videos and inject temporal context through test-time optimization. However, these approaches heavily hinge on precise camera poses and require a long optimization time.

Recent advancements in video generative models~\citep{blattmann2023align, svd, ho2022video, chen2024diffusion} have demonstrated significant growth, achieving high spatial quality and temporal consistency.
We argue video diffusion models can be repurposed for video depth estimation in the wake of diffusion-based single-image depth estimatiors~\citep{Ke2023repurposing,fu2024geowizard}, yielding feed-forward estimation without requiring camera poses across frames.
This repurposing process, however, is not merely a straightforward extension of monocular depth prediction. It requires careful consideration of the contextual information between clips to ensure temporal consistency across a long video.

Therefore, we propose a novel video depth estimator, named \method{}, to leverage a video diffusion model to provide context cues within and between clips, resulting in much higher temporal consistency compared to the use of single-image diffusion models -- see \cref{fig:teaser}, as highlighted by the smoother depth depicted by extracting y-t slices from the results by our model against a state-of-the-art single-image model~\citep{Ke2023repurposing}. 
Specifically, to effectively model context within a single clip with good generalization, we finetune video diffusion models, drawing inspiration from recent efforts that leverage image generative model priors for single-frame tasks~\citep{ji2023ddp, duan2023diffusiondepth, saxena2023ddvm, zhao2023unleashing, du2023generative, Ke2023repurposing, fu2024geowizard, gui2024depthfm, xu2024diffusion}.
In terms of cross-clip context, we systematically evaluate different strategies, and propose a consistent context-aware inference as shown in~\cref{fig:infer_comp}.
At inference time, processing video through a sliding window without carrying context across frames (\cref{fig:infer_comp} (a)) does not ensure temporal consistency. A common method to add context between clips is the ``replacement trick'' in video diffusion models~\cite{ho2022video,opensora}, used in our concurrent work DepthCrafter~\citep{hu2024depthcrafter}. This approach involves initializing overlapping frames by adding noise to previously predicted depth frames (\cref{fig:infer_comp} (b)), but this leads to inconsistent contextual information due to noise variations~\citep{ho2022video}. To address this, our proposed strategy uses previously predicted depth frames for context without adding noise, ensuring consistency across frames (\cref{fig:infer_comp} (c)). This is enabled by applying different noise levels to each frame during training, allowing the model to denoise across varying noise levels.

Furthermore, we conduct thorough comparisons between various protocols to train \method, empirically finding an effective training strategy fully exploring image and video depth datasets, %
identifying a sequential training of spatial and temporal layers as the most effective one, where the spatial layers are first trained and kept frozen during the training of the temporal layers.

Quantitative and qualitative results on open-world video benchmarks confirm that \method{} achieves state-of-the-art temporal consistency on video depth estimation, while maintaining spatial accuracy comparable with state-of-the-art single-image depth estimation methods. 

\section{Related Work}
\label{related_work}

\hspace{0.4cm}\textbf{Discriminative Monocular Depth Estimation.} 
These approaches are trained end-to-end to regress depth, according to two alternative categories: metric versus relative depth estimation. 
Early attempts focused on the former category~\citep{eigen2014depth}, yet were limited to single scenarios --  i.e., training specialized models for driving environments~\citep{Geiger2012CVPR} or indoor scenes~\citep{Silberman2012ECCV}. Among these, some frameworks used ordinal regression~\citep{FuCVPR18}, local planar guidance~\citep{lee2019big}, adaptive bins~\citep{bhat2021adabins} or fully-connected CRFs~\citep{yuan2022newcrfs}.
In pursuit of achieving generalizability across different environments~\citep{ranftl2020towards}, the community recently shifted towards training models estimating relative depth, through the use of affine-invariant loss functions~\citep{ranftl2020towards,ranftl2021vision} to be robust against different scale and shift factors across diverse datasets. On this track, the following efforts focused on recovering the real 3D shape of scenes~\citep{yin2021learning}, exploiting surface normals as supervision~\citep{yin2021virtual}, improving high-frequency details~\citep{Miangoleh2021Boosting,zhang2022hierarchical} or exploiting procedural data generation~\citep{eftekhar2021omnidata}.
To combine the best of the two worlds, new attempts to learn generalizable, metric depth estimation have emerged lately~\citep{bhat2023zoedepth,li2023patchfusion}, either explicitly handling camera parameters through canonical transformations~\citep{yin2023metric,hu2024metric3dv2} or as direct prompts to the model~\citep{tri-zerodepth,piccinelli2024unidepth}.

\textbf{Generative Monocular Depth Estimation.}
Eventually, some methodologies 
have embraced the use of pre-trained generative models for depth estimation. Some exploited Low-Rank Adaptation (LoRA)~\citep{du2023generative} or started from self-supervised pre-training~\citep{saxena2023ddvm}, while others re-purposed Latent Diffusion Models (LDM) by fine-tuning the pre-trained UNet~\citep{Ke2023repurposing} to denoise depth maps. Further advances of this latter strategy jointly tackled depth estimation and surface normal prediction~\citep{fu2024geowizard}, or exploited Flow Matching~\citep{gui2024depthfm} for higher efficiency.
Despite the strong priors learned by diffusion models allow for capturing intricate details in complex environments, these frameworks expose scarce temporal consistency and flickering artifacts over frame sequences.

\textbf{Video Depth Estimation.}
In addition to spatial accuracy, temporal consistency is paramount when predicting depth from videos. %
This task has been mostly tackled through discriminative methods: some approaches estimate the poses of any frame in the video and use them to build cost volumes~\citep{watson2021temporal, long2021multi, guizilini2022multi} or running test-time training~\citep{luo2020consistent, kopf2021robust, zhang2021consistent}, with both heavily depending on the accuracy of the poses and the latter lacking generalizability as it overfits a single video. Others deploy recurrent networks~\citep{zhang2019exploiting,reccurent_depth_2020}, while most recent works exploit attention mechanisms~\citep{wang2022less, cao2021learning}, yet with sub-optimal results compared to state-of-the-art single-image depth predictors. NVDS~\citep{wang2023neural} introduces a stabilization network so as to conduct temporal refinement to single-frame results from an off-the-shelf depth predictor. %
Concurrently with us, DepthCrafter~\cite{hu2024depthcrafter} proposes a generative framework for video depth estimation, using latent interpolation based on the replacement trick to enforce temporal consistency over long clips, leading to inconsistency due to noise variation. %

\textbf{Diffusion Models.}
Image Diffusion Models (IDMs) by Sohl-Dickstein et al.~\citep{sohl2015deep} 
conquered the main stage for image generation tasks~\citep{dhariwal2021diffusion, kingma2021variational} at the expense of GANs. Further developments aimed at improving both generation conditioning~\citep{alembics-disco-diffusion, nichol2021glide} and computational efficiency~\citep{rombach2022high}, with Latent Diffusion Models (LDMs)~\citep{rombach2022high} notably emerging as a solution for the latter. %
Among conditioning techniques, the use of cross-attention~\citep{avrahami2022blended, brooks2023instructpix2pix, gafni2022make, hertz2022prompt, kawar2023imagic, kim2022diffusionclip, nichol2021glide, parmar2023zero, ramesh2022hierarchical} and the encoding of segmentation masks into tokens~\citep{avrahami2023spatext, gafni2022make} stand as the most popular, with additional schemes being proposed to enable the generation of visual data conditioned by diverse factors such as text, images, semantic maps, sketches, and other representations~\citep{bar2023multidiffusion, bashkirova2023masksketch, huang2023composer, mou2023t2i, zhang2023adding, voynov2023sketch}.
Lately, DMs have been extended for video generation~\citep{ho2022video,svd,chai2023stablevideo,jain2024peekaboo,zhang2024trip,wu2023tune}, focusing on obtaining consistent content over time thanks to the integration of temporal layers, incorporating self-attention mechanisms and convolutions between frames within conventional image generation frameworks. One line of work~\cite{ho2022video,chen2024diffusion,gao2024vista} explores infinite video generation via adapting both training and sampling paradigms based on the original diffusion model.
Diffusion Forcing~\cite{chen2024diffusion} is mostly similar with \method{}, yet with notable differences as it
focuses on generation task while \method{} focuses on video depth estimation, a deterministic task.
Furthermore, Diffusion Forcing is instantiated with causal architectures and implemented with a Recurrent Neural Network (RNN), where they maintain hidden states to capture the information of past tokens. However, \method{} is implemented with an attention network without additional hidden states, directly fusing the information of past tokens with current noisy input.

\section{Method}
\label{method}

\newcommand{\rgb}{\mathbf{x}}
\newcommand{\depth}{\mathbf{d}}
\newcommand{\latent}{\mathbf{z}}
\newcommand{\latentdepth}{\latent^{(\depth)}}
\newcommand{\pseudolatent}{\hat{\latent}}
\newcommand{\pseudolatentdepth}{\pseudolatent^{(\depth)}}
\newcommand{\latentrgb}{\latent^{(\rgb)}}
\newcommand{\noise}{\bm{\epsilon}}
\newcommand{\denoiser}{\mathbf{s}_\mathbf{\theta}}
\newcommand{\seconddenoiser}{D_\mathbf{\theta}}
\newcommand{\thirddenoiser}{F_\mathbf{\theta}}
\newcommand{\denoiserlong}{\seconddenoiser(\latentdepth_t, \latentrgb, t)}

\begin{figure*}[t!]
  \centering
  \includegraphics[width=0.95\linewidth]{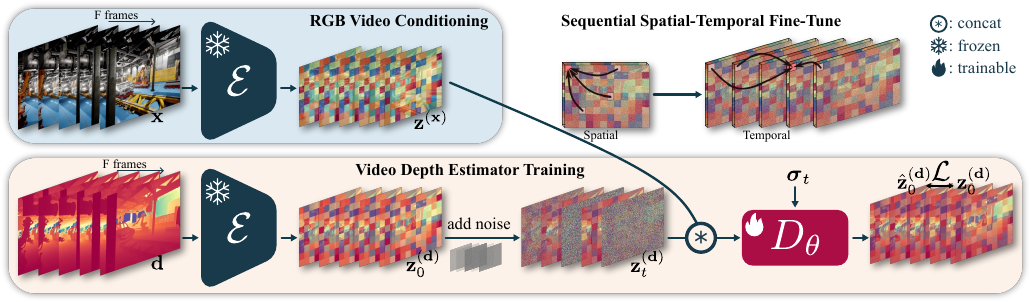}\vspace{-0.3cm}
  \caption{\textbf{Our adapted training pipeline}. %
  We add an RGB video conditioning branch to a pre-trained video diffusion model (SVD) and fine-tune it via DSM for depth estimation, by sampling different noise levels for each frame. %
  Our training involves two stages: 1) we train the spatial layers with single-frame depths; 2) we freeze spatial layers and train the temporal layers on clips of random lengths. %
  }
    \label{fig:training_pipeline}
\vspace{-0.3cm}
\end{figure*}

In this section, we introduce \method{}, a consistent video depth estimator derived from a video foundation model, specifically from Stable Video Diffusion (SVD)~\citep{svd}.
Given an arbitrary-length video, our goal is to generate spatial-accurate and temporal-consistent video depth. 
We first introduce the straightforward way of repurposing SVD for video depth estimation in \cref{sec:naive_formulation} and discuss its issues in artitray-length video inference. Next, we elaborate on our consistent context-aware strategy in \cref{sec:context_aware} and our training protocols in \cref{sec:finetuning}.

\subsection{Naive Diffusion Formulation}
\label{sec:naive_formulation}
In order to align with the video foundation model, SVD~\citep{svd}, a straightforward way is to follow~\citep{Ke2023repurposing} and reformulate monocular video depth estimation as a continuous-time denoising diffusion~\citep{song2020score, karras2022elucidating} generation task conditioned on the RGB video.
The diffusion model consists of a stochastic \textit{forward} pass to inject one noise level Gaussian noise into the input sequence and a \textit{reverse} process to remove noise with a learnable denoiser $\seconddenoiser$.
Following SVD, our diffusion model is defined in a latent space of lower spatial dimension for better computation efficiency, where a variational autoencoder (VAE) consists of an encoder $\cE$ and a decoder $\mathcal{D}$ is used to compress the original signals.
In order to repurpose the VAE of SVD, which accepts a 3-channel (RGB) input, we replicate the depth map into three channels to mimic an RGB image.
Then, during inference, we decode and calculate the average of these three channels, which serves as our predicted depth map following~\citep{Ke2023repurposing}.

\noindent\textbf{Training.} 
During training, one $F$-frame RGB video clip $\rgb \in \mathbb{R}^{F \times W\times H \times 3}$ and the corresponding depth $\depth \in \mathbb{R}^{F \times W\times H }$ are first encoded into the latent space with the VAE encoder: $\latentrgb=\cE(\rgb), \latentdepth_0=\cE(\depth)$. 
For each training step, we sample one noise level $\sigma_t$ for the whole clip , where $\log\sigma_t \sim \cN(P_{mean}, P_{std}^2)$~\citep{karras2022elucidating}
with $P_{mean}=0.7$ and $P_{std}=1.6$.
Then we add noise with this noise level to the depth latent $\latentdepth_0$ to obtain the noisy depth latent $\latentdepth_t$ as
\begin{equation} 
    \latentdepth_t=\latentdepth_0+\sigma_t^2\boldsymbol{\epsilon}, \ \boldsymbol{\epsilon} \sim \cN(\mathbf{0}, \mathbf{I}).
\end{equation}
In the reverse process, diffusion model denoises $\latentdepth_t$ towards predicted clean $\pseudolatentdepth_0$ with a learnable \textit{denoiser} $\seconddenoiser$ as 
\begin{equation} \label{eq:naive_denoise}
    \pseudolatentdepth_0=D_\theta(\latentdepth_t; \sigma_t, \latentrgb),
\end{equation}
which is trained via \textit{denoising score matching}~(DSM)~\citep{vincent2011connection}
\begin{equation} \label{eq:naive_diffusion_objective}
    \cL=\mathbb{E}_{\latentdepth, \latentrgb, \sigma_t} \left[\lambda(\sigma_t) \| \pseudolatentdepth_0- \latentdepth_0 \|_2^2 \right],
\end{equation}
with weighting function $\lambda(\sigma)=(1+\sigma^2)\sigma^{-2}$.
In this work we follow EDM~\citep{karras2022elucidating} and parametrize the denoiser $\seconddenoiser$ as 
\begin{equation} \label{eq:naive_edm_preconditioning}
\begin{split}
    \seconddenoiser(\latentdepth_t; &\sigma_t, \latentrgb) = 
    c_\text{skip}(\sigma_t)\latentdepth_t + \\ c_\text{out}&(\sigma_t) \thirddenoiser(c_\text{in}(\sigma_t)\latentdepth_t; c_\text{noise}(\sigma_t), \latentrgb),
\end{split}
\end{equation}
where $\thirddenoiser$ is a UNet to be trained in our case, and $c_\text{skip}$, $c_\text{out}$, $c_\text{in}$, and $c_\text{noise}$ are preconditioning functions.
The condition, RGB video latent $\latentrgb$, in this formulation is introduced via concatenation with depth video latent on feature dimension.

\noindent\textbf{Single-Clip Inference.} 
At testing time, single-clip depth latent $\pseudolatentdepth_0$ is restored from a randomly-sampled Gaussian noise $\pseudolatentdepth_T$ conditioning on the RGB clip by iteratively applying the denoising process with trained \textit{denoiser} $\seconddenoiser$ 
\vspace{-0.2cm}
\begin{equation} \label{eq:naive_initialization}
    \pseudolatentdepth_T\sim \cN(\mathbf{0}, \sigma_T^2\mathbf{I})
\end{equation}
\vspace{-0.5cm}
\begin{equation} \label{eq:naive_infer_denoise}
    \tilde{\latent}^{(\depth)}_0=D_\theta(\latentdepth_t; \sigma_t, \latentrgb)
\end{equation}
\vspace{-0.7cm}
\begin{equation} \label{eq:inference}
    \pseudolatentdepth_{t-1} = \frac{\pseudolatentdepth_{t} - \tilde{\latent}^{(\depth)}_0}{\sigma_{t}} (\sigma_{t-1}-\sigma_{t}) + \pseudolatentdepth_{t},  \quad 0 < t \le T
\end{equation}
where $\sigma_0,...,\sigma_T$ are sampled from a fixed variance schedule of a denoising process with $T$ steps. Video depth $\hat{\depth}$ could be obtained with the VAE decoder $\hat{\depth}=\mathcal{D}(\pseudolatentdepth_0)$.

\begin{figure*}[t!]
  \centering
  \includegraphics[width=0.95\linewidth]{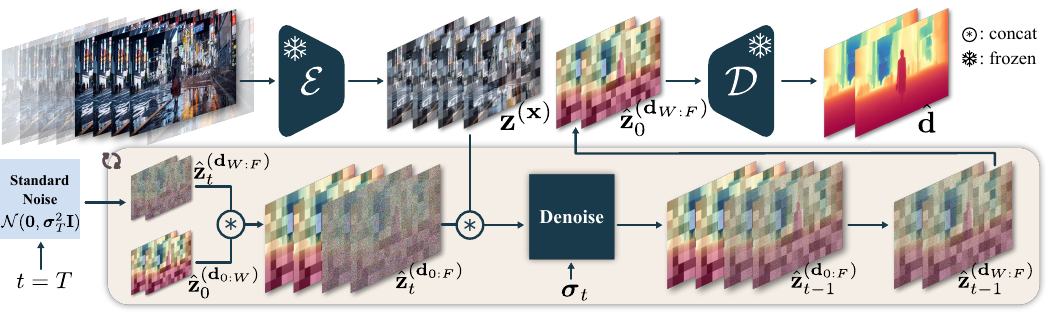}\vspace{-0.3cm}
  \caption{\textbf{Inference pipeline of our consistent context-aware inference strategy}. %
  Given an infinitely long video, we segment it into several $F$-frame clips with overlap $W$ and use a sliding window strategy for inference. Besides, we initialize overlapping $W$ frames with previously predicted depth frames $\pseudolatent^{(\depth_{0:W})}_0$
  to support consistent context information during each sampling step.
                }
    \label{fig:inference_pipeline}
\vspace{-0.3cm}
\end{figure*}

\noindent\textbf{Arbitrary-Length Video Inference.} 
Based on the purpose of autoregressive inference and the limitation of computing resources, it is necessary to split a long video into smaller video clips and process them in streamed manner. 
A solution could consists of using a naive \textit{sliding window} strategy: the given video sequence is split into multiple independent video clips with overlap $W$. The model makes independent inferences for each video clip: the whole clip is initialized via \cref{eq:naive_initialization} and updated via \cref{eq:inference} ( see~\cref{alg:naive_sliding_window}).
The depth latent is sampled from $p_\theta(\pseudolatent^{(\depth_{W:F})}_{t-1}|\pseudolatentdepth_{t},\rgb)$, therefore without any contextual information exchange between clips, causing flickering artifacts.

Inspired by recent advancements in video generative models~\citep{ho2022video}, a straightforward method to share contextual information across clips involves conditioning via \textit{replacement trick} -- i.e., employing a sliding window strategy and initializing overlapping frames by adding noise to previously predicted depth frames. Let $\pseudolatent^{(\depth_{0:W})}_0$ denote the latent code of $W$ previously predicted depth frames.
The first $W$ frames are obtained via forward diffuse as below
\begin{equation}
\begin{split}
    \pseudolatent^{(\depth_{0:W})}_t = \pseudolatent^{(\depth_{0:W})}_0+\sigma_t \boldsymbol{\epsilon}, \ \boldsymbol{\epsilon} \sim \cN(\mathbf{0}, \mathbf{I}),  \ 0 < t \le T,
\end{split}
\end{equation}
and the last $F-W$ frames are initialized via \cref{eq:naive_initialization} and updated via \cref{eq:inference}, 
where $\pseudolatentdepth_t = \left [\pseudolatent^{(\depth_{0:W})}_t, \pseudolatent^{(\depth_{W:F})}_t \right ]$ (see~\cref{alg:replacement_trick}).
In this way, the content is injected to some extent. However, \citet{ho2022video} highlights that this conditioning approach lacks mathematical rigor and results in inconsistent contextual information during diffusion sampling at each step (see~\cref{sec:supp_math}). 
The depth latent currently is sampling from $p_\theta(\pseudolatent^{(\depth_{W:F})}_{t-1}|\pseudolatent^{(\depth_{0:W})}_{t}, \pseudolatent^{(\depth_{W:F})}_{t}, \rgb)$. 
Due to the noise added to $\pseudolatent^{(\depth_{0:W})}_{t}$, the conditioning varies during diffusion sampling at each step 
causing discrepancies in the generated sequence. Consequently, when this trick is applied to video depth tasks, it exacerbates inconsistency across clips, manifesting both on geometry and scale.

\subsection{Consistent Context-Aware Strategy}
\label{sec:context_aware}
The issue with the replacement trick is that, at each sampling step, the conditioning changes due to the varying noise added each time, leading to fluctuating guidance. 
Ideally, the previous inference results should be provided without noise to guide the prediction of remaining frames.
Based on this intuition, we propose a novel consistent context-aware strategy, inspired by~\cite{chen2024diffusion}. During training, instead of sampling a single noise level for the entire video clip, we independently sample distinct noise levels for each individual frame within the clip. During inference, we initialize overlapping frames with previously predicted depth and adjust the conditioning noise levels to each frame.

\noindent\textbf{Training.} 
We independently sample distinct noise levels for each individual frame within the clip as
$\boldsymbol{\sigma}_t$
, where $\boldsymbol{\sigma}_t=[\sigma_1, \sigma_2, \dots, \sigma_F]$, $\log\sigma_i \sim \cN(P_{mean}, P_{std}^2)$ 
for $i=1,2,\dots,F$. Then we add these F noise levels to the depth latent $\latentdepth_0$ to obtain the noisy depth latent $\latentdepth_t$ as
\begin{equation} 
    \latentdepth_t=\latentdepth_0+\boldsymbol{\sigma}_t^2\boldsymbol{\epsilon}, \ \boldsymbol{\epsilon} \sim \cN(\mathbf{0}, \mathbf{I}).
\end{equation}
In the reverse process, we correspondingly adapt the previous \cref{eq:naive_diffusion_objective,eq:naive_denoise,eq:naive_edm_preconditioning} as follows
\begin{equation}
    \pseudolatentdepth_0=D_\theta(\latentdepth_t; \boldsymbol{\sigma}_t, \latentrgb),
\end{equation}
\vspace{-0.5cm}
\begin{equation} \label{eq:our_diffusion_objective}
    \cL=\mathbb{E}_{{\latentdepth, \latentrgb, \boldsymbol{\sigma}_t}} \left[\lambda(\boldsymbol{\sigma}_t) \| \pseudolatentdepth_0- \latentdepth_0 \|_2^2 \right],
\end{equation}
\vspace{-0.5cm}
\begin{equation} \label{eq:our_edm_preconditioning}
\begin{split}
    \seconddenoiser(\latentdepth_t; &\boldsymbol{\sigma}_t, \latentrgb) = 
    c_\text{skip}(\boldsymbol{\sigma}_t)\latentdepth_t + \\ c_\text{out}&(\boldsymbol{\sigma}_t) \thirddenoiser(c_\text{in}(\boldsymbol{\sigma}_t)\latentdepth_t; c_\text{noise}(\boldsymbol{\sigma}_t), \latentrgb).
\end{split}
\end{equation}
\cref{fig:training_pipeline} contains an overview of our adapted training pipeline.

\noindent\textbf{Arbitrary-length video inference.} 
There is no previously predicted depth available for the first clip, though it is available for the subsequent clips. 
Therefore we perform inference of the first clip using the standard inference strategy as~\cref{eq:naive_initialization,eq:naive_infer_denoise,eq:inference}. 
For subsequent clips, we initialize overlapping $W$ frames with previously predicted depth frames $\pseudolatent^{(\depth_{0:W})}_0$ without adding any noise, and the last $F-W$ frames are initialized via \cref{eq:naive_initialization}. 
Note that the noise condition of the denoiser $D_\theta$ is changed correspondingly as
$\boldsymbol{\sigma}_t = [\underbrace{\sigma_{\epsilon}, \sigma_{\epsilon}, ..., \sigma_{\epsilon}}_{W}, \underbrace{\sigma_{t}, \sigma_{t}, ..., \sigma_{t}}_{F-W}]$
and $\sigma_{\epsilon}$ indicates a very small noise level (see~\cref{alg:ours} for pseudocode).
Training with our adapted pipeline has empowered our model to effectively denoise different frames within one clip under different noise levels. 
In this way, the depth latent is sampling from $p_\theta(\pseudolatent^{(\depth_{W:F})}_{t-1}|\pseudolatent^{(\depth_{0:W})}_0, \pseudolatent^{(\depth_{W:F})}_{t}, \rgb)$.
This method ensures that the contextual information between clips remains consistent across any sampling step. 
Moreover, the rationale behind conditioning previously predicted depth frames with a small noise level rather than a clean noise level is that such depth frames are not ground-truth, so they cannot be fully trusted. 
This small noise level accounts for the inherent uncertainty from the previous inference and mitigates long-term compounding errors, thereby enhancing the robustness of our model. We provide experimental results of this small noise level in~\cref{sec:add_ablation}. This idea is also explored in a concurrent work, Diffusion Forcing~\citep{chen2024diffusion}, focusing on generative modeling, whereas we uncover its critical role in generative-based video depth estimation. 
Our overall inference strategy is depicted in \cref{fig:inference_pipeline}. 

\begin{figure*}[t!]
  \begin{minipage}[t]{0.29\linewidth}
    \centering
    \footnotesize
    \input{algos/naive_sliding_window}
  \end{minipage}
  \hfill
  \begin{minipage}[t]{0.32\linewidth}
    \centering
    \footnotesize
    \input{algos/replacement_trick}
  \end{minipage}
  \begin{minipage}[t]{0.38\linewidth}
    \centering
    \footnotesize
    \input{algos/ours}
  \end{minipage}
  \vspace{-12pt}
\end{figure*}

\subsection{Multi-Stage Training Protocol}
\label{sec:finetuning}
In contrast to image-generative models, video-generative models incorporate both spatial and temporal dimensions. 
One central question thereby becomes how to effectively fine-tune these pre-trained video generative models to achieve a satisfying level of spatial accuracy and temporal coherence in geometry.
In response to this, we have conducted comprehensive analyses to ascertain the best practices for taming the initial foundational video model into a consistent depth estimator.
Note that the VAE for compressing the original pixel space remains frozen.

\noindent\textbf{Using Single-Frame Datasets.} 
The availability of single-frame depth datasets is generally higher than the one of video depth datasets. These latter usually count more frames compared to the former, yet they feature a reduced number of scenes and thus lack diversity.
Accordingly, we argue that jointly using single-frame and multi-frame depth datasets can play a significant role in achieving good spatial and temporal accuracy. 
Therefore, we also make use of the single-frame datasets throughout the full training process, in contrast to SVD~\citep{svd} that uses single-view image datasets~\citep{schuhmann2022laion} only during the first pre-training stage.

\noindent\textbf{Randomly Sampled Clip Length.} 
As introduced before, splitting the video into fixed-length clips could be a natural choice at inference time to maintain consistent resource requirements. However, a fixed-length clip could involve slower or faster motion, making it harder for the model to generalize. 
Accordingly, we argue that sampling clips of random length at training time can act as a form of data augmentation, making the model more robust to such different behaviors. Both fixed and random length choices are naturally supported by design, thanks to the spatial layers interpreting the video as a batch of independent images and temporal layers applying both convolutions and temporal attention along the time axis, thus allowing us to randomly sample clips of length $F \in [1, F_{\text{max}}]$ during training. %

\noindent\textbf{Sequential Spatial-Temporal Fine-Tuning.} 
The original SVD fine-tunes the full UNet $D_\theta$ on video datasets, meaning that both the spatial layers and the temporal layers are trained jointly. 
However, we argue splitting the fine-tuning of spatial and temporal components into distinct phases could favor achieving the highest temporal consistency that we pursue when dealing with videos. 
Accordingly, we investigate an alternative training protocol -- sequential spatial-temporal training. Specifically, we first train the spatial layers using single-frame supervision. After convergence, we keep the spatial layers frozen and fine-tune the temporal layers using clips of randomly sampled length as supervision. We will demonstrate how this sequential protocol favors temporal consistency while maintaining spatial accuracy intact.

\begin{table*}[t!]
    \centering
    \renewcommand{\tabcolsep}{6pt}
    \resizebox{0.95\linewidth}{!}{
    \begin{tabular}{
    @{}l|
    ccc|
    ccc|
    cccc||
    ccc 
    c@{}
    }
    
    \toprule
    \multirow{2}{*}{Method} &  
    \multicolumn{3}{c|}{KITTI-360} & 
    \multicolumn{3}{c|}{ScanNet++} &
    \multicolumn{4}{c||}{Sintel} &
    
    \multicolumn{3}{c}{Avg. Rank $\downarrow$} \\
    
    & 
    
    AbsRel$\downarrow$ & $\delta$1$\uparrow$ & \textit{MFC}$\downarrow$  & 
    
    AbsRel$\downarrow$ & $\delta$1$\uparrow$ & \textit{MFC}$\downarrow$  & 

    AbsRel$\downarrow$ & $\delta$1$\uparrow$ & \textit{MFC}$\downarrow$ & \textit{MFC}$^* \downarrow$ &

    AbsRel & $\delta$1 & \textit{MFC} 
    
    \\
    
    \midrule
    
    Marigold~\citep{Ke2023repurposing} &
    
    \snd{0.213} & 
    \bf\fst{0.665} &
    0.776 & 
     
    0.192 & 
    0.699 &
    0.109 & 
    
    0.573 & 
    0.529 &
    \trd{1.112} &
    0.932 & 

    4.33 &
    3.33 &
    4.00 \\

    Marigold (SVD) &
    
    0.247 & 
    0.608 &
    \trd{0.694} & 
     
    0.197 & 
    0.686 &
    0.112 & 
    
    0.539 & 
    0.510 &
    \snd{1.005} &
    \snd{0.796} & 

    5.33 &
    5.33 &
    \trd{3.67} \\

    DepthAnything~\citep{depthanything} &
    
    0.215 & 
    0.635 &
    0.952 &

    \bf\fst{0.170} & 
    \trd{0.712} &
    \trd{0.103} & 
    
    \bf\fst{0.329} & 
    \snd{0.565} &
    1.399 &
    1.038 &

    \bf\fst{1.00} &
    \trd{3.00} &
    5.00 \\

    DepthAnything V2~\citep{yang2024depth} &

    \bf\fst{0.207} & 
    \snd{0.656} &
    0.807 &
    
    \bf\fst{0.170} & 
    \snd{0.713} &
    \trd{0.103} & 
    
    \trd{0.387} & 
    0.554 &
    1.504 &
    1.125 &

    \snd{1.67} &
    \snd{2.67} &
    5.00 \\
    
    \midrule
    NVDS~\citep{wang2023neural} &
    
    0.379 & 
    0.384 &
    1.276 &
    
    0.239 & 
    0.565 &
    0.136 & 
    
    0.442 & 
    0.465 &
    1.220 &
    0.924 &

    6.00 &
    7.00 &
    6.00 \\
    
    DepthCrafter~\citep{hu2024depthcrafter} &
     
    0.293 & 
    0.462 &
    \snd{0.655} &
    
    0.199 & 
    0.642 &
    \snd{0.094} & 
    
    \snd{0.374} & 
    \bf\fst{0.566} &
    1.270 &
    \trd{0.889} &

    4.67 &
    4.33 &
    \snd{3.00} \\

    \bf\method{} (Ours) &
     
    \trd{0.215} & 
    \trd{0.654} &
    \bf\fst{0.407} &
    
    \trd{0.176} & 
    \bf\fst{0.726} &
    \bf\fst{0.092} & 
    
    0.493 & 
    \trd{0.555} &
    \bf\fst{0.728} &
    \bf\fst{0.516} &

    \trd{4.00} &
    \bf\fst{2.33} &
    \bf\fst{1.00} \\
    
    \bottomrule

    \end{tabular}}
    \vspace{-0.3cm}
    \caption{{\bf Quantitative Comparison on zero-shot depth benchmarks.} Top: single-image depth estimators. Bottom: video depth estimators. \textit{MFC}$^*$ denotes the use of ground-truth optical flow to account for moving objects when computing \textit{MFC}.  
    }\vspace{-0.4cm}
\label{tab:comp_main}

\end{table*}

\section{Experimental Results}
\label{experiment}

\subsection{Training Protocol}
\noindent\textbf{Implementation Details.}
We implement \method{} using diffusers by fine-tuning SVD, specifically the image-to-video variant, following the strategy discussed in \cref{sec:finetuning}. 
We disable the cross-attention conditioning of the original SVD, 
we use the standard EDM noise schedule and the network preconditioning~\citep{karras2022elucidating} and set image and video resolution to $576 \times 768$. 
We first pre-train our model with single-frame samples as input for 20k steps, using batch size 8; then, we fine-tune it with F-frame video clips for 18k steps using batch size 1, with F randomly sampled in $[1, F_\text{max}]$, setting $F_\text{max} = 5$. %
For the consistent context-aware inference strategy, we set the length of each clip $T=10$ and the number of overlapping frames between two adjacent clips $W = 5$. %
The entire training takes about 1.5 days on a cluster of 8 Nvidia Tesla A100-80GB GPUs. 
We use the Adam optimizer with a learning rate of 3e-5.

\noindent\textbf{Training Datasets.}
We utilize four synthetic datasets, including single-frame and multi-frame datasets. 
For the former category, we use \textit{Hypersim}~\citep{roberts2021hypersim}, a photorealistic synthetic dataset with 461 indoor scenes. We use the official split with around 54K samples from 365 scenes for training. We filter out incomplete samples, yielding 39K samples.
The latter category involves multiple datasets: 
\textit{Tartanair}~\citep{wang2020tartanair} is a dataset collected in simulation environments for robot navigation tasks including 18 scenes. We use all the 738 sequences for training. 
\textit{Virtual KITTI 2}~\citep{cabon2020virtual} is a synthetic urban dataset providing 5 scenes with various weather or modified camera configurations. We use 4 scenes with 80 sequences for training. 
\textit{MVS-Synth}~\citep{huang2018deepmvs} is a synthetic urban dataset captured in the video game GTA. We use all 120 sequences for training. 
In total, our training set counts 39K single-frame samples and 938 video sequences.

\subsection{Evaluation Protocol}
\label{subsection:evaluation}

\begin{figure*}[t!]
  \centering
  \includegraphics[width=0.90\linewidth]{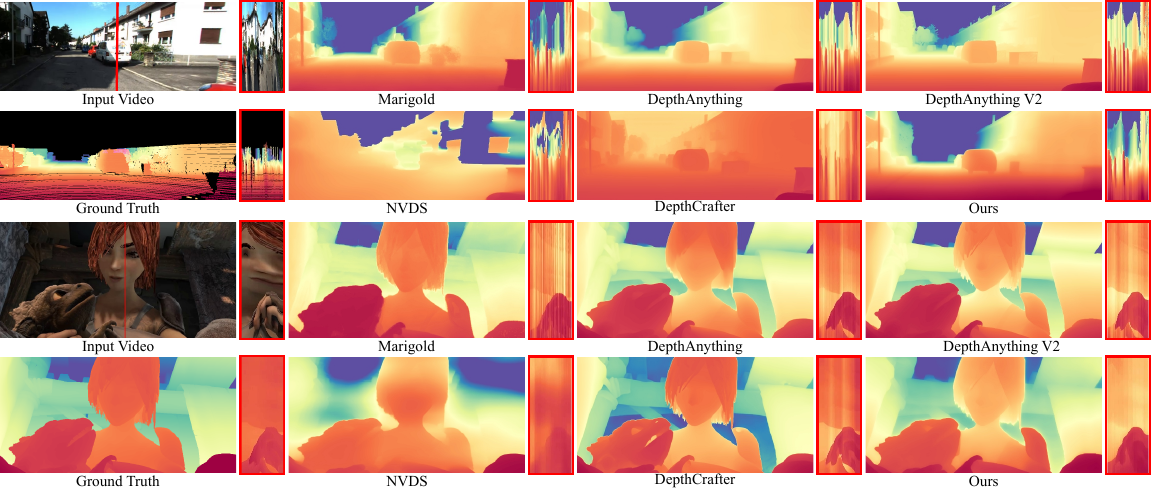}
  \vspace{-0.3cm}
      \caption{\textbf{Qualitative comparison for video depth estimation.} For enhanced visualization of temporal quality, we present the y-t slice of each result within red boxes, achieved by slicing the depth values along the time axis at the designated red line positions.}
    \label{fig:qualitative}
\vspace{-0.4cm}
\end{figure*}

\noindent\textbf{Datasets.}
We quantitatively evaluate our model on three video datasets, ranging from synthetic to realistic, indoor to outdoor, and static to dynamic, thereby assessing the generalization ability of our model across diverse open-world scenarios: %
\textit{KITTI-360}~\citep{kitti360} is a driving dataset collected in urban scenes, from which we select 8 video sequences with 200 frames each. We project LiDAR point clouds into image space to obtain sparse ground-truth depth. 
\textit{ScanNet++}~\citep{yeshwanthliu2023scannetpp} is an indoor dataset featuring depth maps captured by a laser scanner. 
In our evaluation, we use the \textit{nvs\_sem\_val} set, which comprises 50 video sequences across diverse scenes from which we selected the first 90 frames for each.
\textit{Sintel}~\citep{butler2012naturalistic} is a synthetic dataset with precise depth labels.
It comprises 23 sequences, each approximately 50 frames in length, within the training set.

\noindent\textbf{Metrics.}
Following the protocol for affine-invariant depth evaluation~\citep{ranftl2020towards}, we initially align the estimated depth with ground-truth depth using least squares fitting. 
Differently from single-image depth evaluation, for which a different pair of scale and shift factors is fitted for each frame, a global pair of scale and shift values is estimated over the entire video following \cite{godard2019digging,zhang2021consistent} to reflect scale consistency.
Then, we consider two widely recognized metrics~\citep{ranftl2020towards} for evaluation, Absolute Mean Relative Error (\textit{AbsRel}) and $\delta 1$ accuracy with a specified threshold value of 1.25. 
To measure \textit{temporal consistency}, we introduce multi-frame consistency (\textit{MFC}) by warping the prediction from one frame to its adjacent frame, and evaluating the discrepancy between the two depth map predictions.
On datasets featuring several moving objects such as Sintel, we use ground-truth optical flow to compute \textit{MFC}.
Note that both ground-truth flow and camera poses are used only to compute \textit{MFC}. Please refer to~\cref{sec:supp_eval} for details of the \textit{MFC} metric.

\begin{table*}[t!]
    \centering
    \renewcommand{\tabcolsep}{10pt}
    \resizebox{0.92\linewidth}{!}{
    \begin{tabular}{cccc|ccc|ccc|ccc}
    \toprule
    &\multicolumn{3}{c|}{Training} & \multicolumn{3}{c|}{Inference} & \multicolumn{3}{c|}{KITTI-360} & \multicolumn{3}{c}{ScanNet++}  
    \\
    &Image &
    RandomClip &
    S-T FT &
    Naive &
    Replacement &
    Ours & 
    \textit{AbsRel}$\downarrow$ & $\delta$1$\uparrow$ & 
    \textit{MFC}$\downarrow$ &
    \textit{AbsRel}$\downarrow$ & $\delta$1$\uparrow$ &
    \textit{MFC}$\downarrow$ \\ 
    \midrule
    (A) & & \checkmark & \checkmark & 
    \checkmark &  &  &  
    0.252 & 0.575 & 0.555 & 0.201 & 0.676 & \snd{0.097}
     \\
    (B)&\checkmark &  & \checkmark & 
    \checkmark &  &  &  
    0.236 & 0.609 & \snd{0.470} & 0.205 & 0.669 & 0.104     
     \\
    (C)&\checkmark & \checkmark &  & 
    \checkmark &  &  &  
    \snd{0.229} & \snd{0.625} & 0.482 & 0.224 & 0.631 & 0.107
    \\
    \midrule
    (D)&\checkmark & \checkmark & \checkmark & 
    \checkmark &  &  &  
    0.233 & 0.614 & 0.505 & \snd{0.194} & \trd{0.692} & 0.098 
    \\
    (E)&\checkmark & \checkmark & \checkmark & 
    & \checkmark &  &  
    \trd{0.231} & \trd{0.618} & \trd{0.479} & \snd{0.194} & \snd{0.693} & \snd{0.097}
    \\
    (F)&\checkmark & \checkmark & \checkmark & 
    &  & \checkmark &  
    \bf\fst{0.215} & \bf\fst{0.654} & \bf\fst{0.407} & \bf\fst{0.176} & \bf\fst{0.726} & \bf\fst{0.092}
    \\
    \bottomrule
    \end{tabular}}\vspace{-0.3cm}
    \caption{{\bf Quantitative ablation studies} We measure the impact of different pre-training protocols and inference strategies.
    }\vspace{-0.3cm}
    \label{tab:ablation}
\end{table*}

\subsection{Comparison with State-of-the-Art}
\noindent\textbf{Baselines.}
We compare \method{} with five existing baselines, covering both single-image and video depth estimation. %
For single-image depth estimation, we select both generative and discriminative state-of-the-art methods -- respectively Marigold~\citep{Ke2023repurposing} versus DepthAnything~\citep{depthanything} and DepthAnything V2~\citep{yang2024depth}. %
Additionally, we employ an improved Marigold model derived by repurposing SVD in place of Stable Diffusion and deploy it for single-image depth estimation -- Marigold (SVD).
For video depth estimation, we select a discriminative video depth estimator, NVDS~\citep{wang2023neural}, and DepthCrafter~\citep{hu2024depthcrafter}, a video depth estimator derived from SVD~\citep{svd} concurrently with \method{}. We employ the default paper settings for any method, except for DepthCrafter. 
The primary reason for this deviation is that DepthCrafter conditions the prediction process over all frames in a video sequence, thus implying that all frames are known in advance -- which is %
not available when processing videos in a streamed manner through autoregressive prediction. Accordingly, we set the overlap and length of the clip to 5 and 10 to align it with \method{}.

\noindent\textbf{Quantitative Analysis.}
In \tabref{tab:comp_main}, we compare our model with both existing single-image (top) and video (bottom) depth estimators. We highlight the \colorbox{firstcolor}{best}, \colorbox{secondcolor}{second-best}, and \colorbox{thirdcolor}{third-best} scores achieved on any metrics, and report the average ranking achieved across the three datasets on the rightmost columns.
Starting with KITTI-360 and ScanNet++, 
on the one hand, \method{} retains spatial accuracy comparable to state-of-the-art single-frame depth estimators such as Depth Anything and Depth Anything V2, despite these latter being trained on about 500$\times$ the number of images we used to train \method{}; on the other hand, \method{} achieves state-of-the-art temporal consistency -- see the \textit{MFC} metric, with a relative improvement of 98\% on KITTI-360. 
Compared to other video methods such as NVDS and the concurrent DepthCrafter, \method{} outperforms them both in terms of spatial accuracy and temporal consistency.
On Sintel, despite the lower performance in terms of \textit{AbsRel}, \method{} excels in $\delta$1 and maintains superior temporal consistency compared to the other baselines, confirming to be better suited for applications like AR/VR where temporal consistency plays a predominant role over spatial accuracy.

\noindent\textbf{Qualitative Analysis.} \cref{fig:qualitative} shows a qualitative comparison between the baselines and \method{} on video depth estimation. %
For easing the perception of temporal consistency, we present the y-t slice of each result within red boxes, extracted by slicing the depth values along the time axis at the designated red line positions following~\citep{zhang2021consistent}. \method{} generates smooth depth over time while preserving fine-grained details. In contrast, single-frame depth estimators display high-frequency bands in their y-t slice, indicative of flickering artifacts in the estimated depth sequences. Additionally, our two video depth competitors also exhibit similar artifacts at a per-clip level.

\subsection{Ablation Studies}
In~\tabref{tab:ablation}, we run ablation studies on the \textit{KITTI-360} and \textit{ScanNet++} datasets, analyzing the following factors:

\begin{figure}[t!]
  \centering
  \includegraphics[width=0.95\columnwidth]{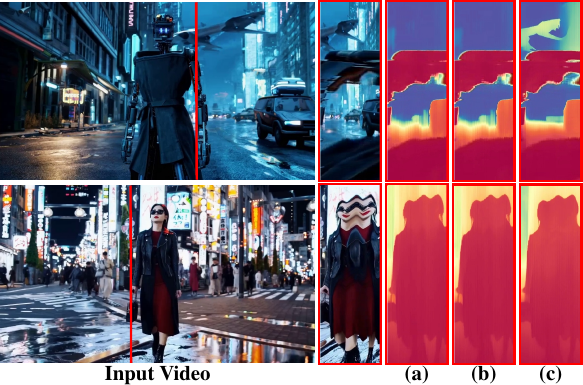}\vspace{-0.3cm}
  \caption{\textbf{
					Qualitative ablation results.} (a) Naive sliding window inference strategy; (b) Inference with replacement trick; (c) Our consistent context-aware inference.
                }
    \label{fig:ablation}
\vspace{-0.3cm}
\end{figure}

\noindent\textbf{Single-Frame Training Datasets.} 
We investigate the impact of single-frame data used during the training phase. %
Our experiments highlight that merely using multi-frame data only (A) yields sub-optimal results over combining both multi-frame and single-frame datasets (D), with this latter improveing spatial accuracy significantly.

\noindent\textbf{Random Clip Length Sampling.} 
Next, we ablate the effectiveness of random video clip length sampling during training.
Removing this sampling (B) leads to performance degradation. This indicates that sampling clips of random length acts as an effective form of data augmentation, mitigating the risk of model overfitting.

\noindent\textbf{Sequential Spatial-Temporal Fine-Tune.} 
We evaluate the effect of sequential spatial-temporal training protocol %
by jointly training the full network (C). %
The sequential training strategy always leads to better temporal consistency, and better spatial accuracy on \textit{ScanNet++}, which means disentangling spatial and temporal layers could be the better way to tame video foundation into a depth estimator. Due to the minimal view change and extensive depth range on \textit{KITTI-360}, both \textit{AbsRel} and $\delta$1 exhibit limited sensitivity.

\noindent\textbf{Inference Strategy.} 
We compare the results obtained by naive sliding window inference (D), %
inference with replacement trick (E) %
and our consistent context-aware inference (F). Compared to naive sliding window inference, inference with replacement trick barely improves the temporal consistency, whereas our approach leads to better results in terms of both spatial accuracy and temporal consistency. We also visualize the y-t slice for these three inference strategies in~\cref{fig:ablation}. Both naive sliding window inference and inference with replacement trick exhibit high-frequency bands at a per-clip level, while our consistent context-aware inference strategy ensures temporal consistency by providing consistent contextual information over the clip, significantly reducing flickering artifacts between windows.

\section{Conclusion}
\label{conclusion}
This paper introduced \method{}, a video depth estimator that prioritizes temporal consistency by leveraging video generation priors. Our exploration of various training and inference strategies has led us to identify the most effective approach to tame SVD into a consistent depth predictor, resulting in superior performance. Specifically, \method{} outperforms existing methods in terms of temporal consistency, surpassing both image and video depth estimation techniques, while maintaining comparable spatial accuracy. We contend our empirical insights into harnessing video generation models for depth estimation lay the groundwork for future investigations in this domain.

\vspace{0.22cm}
{\small
\noindent\textbf{Acknowledgements:}
This work is supported by NSFC under grant U21B2004, 62202418, and 62441223.
This work was supported by Ant Group Research Fund.
Yiyi Liao is with the Zhejiang Provincial Key Laboratory of Information Processing, Communication and Networking (IPCAN).
}

{
    \small
    \bibliographystyle{ieeenat_fullname}
    \bibliography{bibliography_long, bibliography_custom, bibliography}
}

    \clearpage
    \appendix
    \renewcommand*{\thesection}{\Alph{section}}
    \newcommand{\multiref}[2]{\cref{#1}--\ref{#2}}
    \renewcommand{\thetable}{S\arabic{table}}
    \renewcommand{\thefigure}{S\arabic{figure}}
    \setcounter{table}{0}
    \setcounter{figure}{0}
    \twocolumn[{%
    		\renewcommand\twocolumn[1][]{#1}%
                \maketitlesupplementary
                \vspace{-0.3cm}
    		\begin{center}
    			\includegraphics[width=0.95\textwidth]{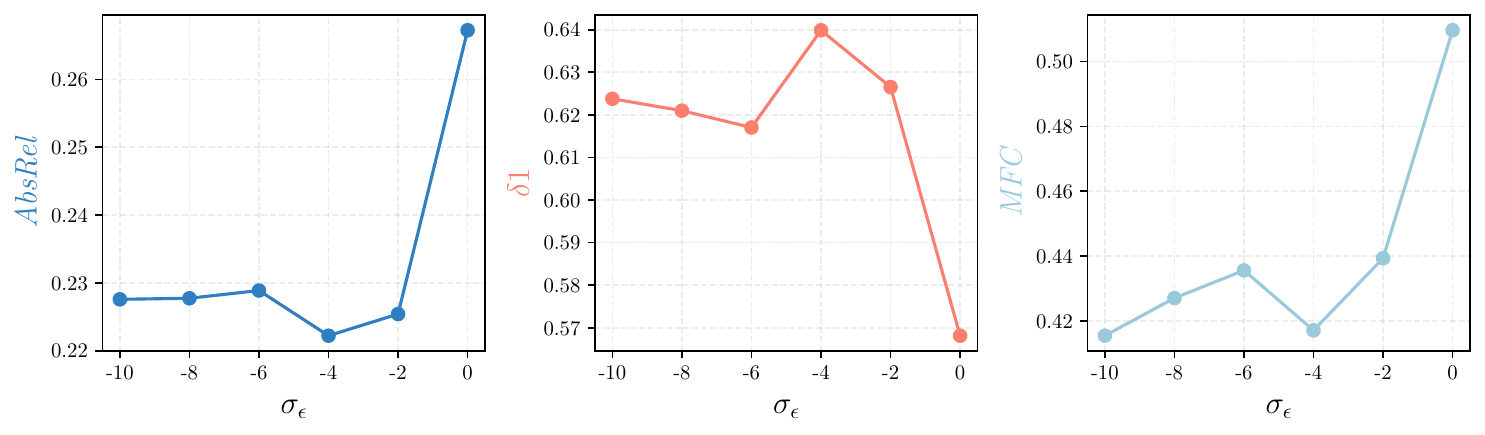}
    \vspace{-0.3cm}			
       \captionsetup{type=figure}
    			\captionof{figure}{
    				\textbf{
    					Ablation Study.} We report accuracy and consistency metrics of our method on \textit{KITTI-360} with different $\sigma_\epsilon$.
    			}
    			\label{fig:ablation_sigma}
    		\end{center}
        \begin{center}
    			\includegraphics[width=0.95\textwidth]{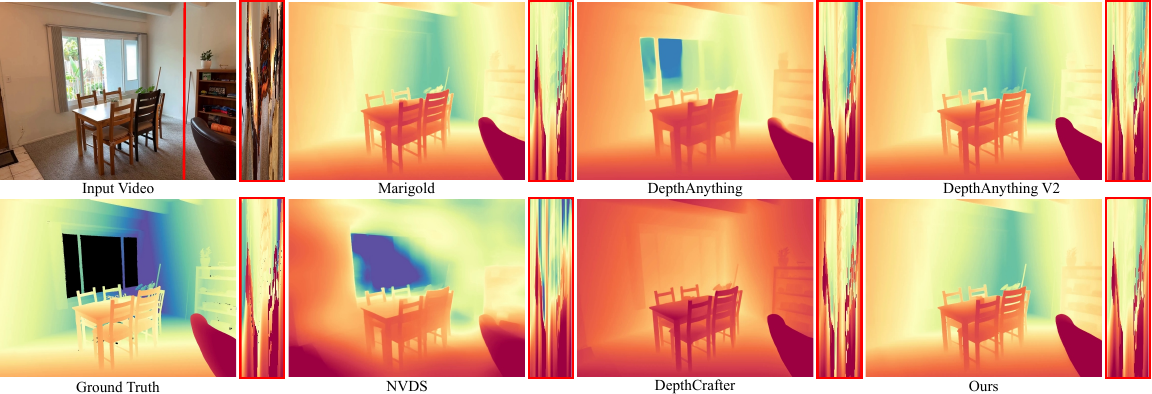}
    \vspace{-0.1cm}			
       \captionsetup{type=figure}
    			\captionof{figure}{
    				\textbf{More qualitative comparison for video depth estimation.} Results on \textit{ScanNet++} dataset.}
    			\label{fig:more_qualitative}
    		\end{center}
    	}]

\setcounter{page}{1}
\section{Details on Evaluation Protocol}
\label{sec:supp_eval}
\noindent\textbf{Datasets.}
The sequences on \textit{KITTI-360} %
we chose follows the rules below. Given the absence of ground-truth poses for the initial frames of each sequence, we extracted frames 300-500, ultimately utilizing 200 frames for evaluation.

\noindent\textbf{Metrics.}
To measure \textit{temporal consistency}, we introduce multi-frame consistency (\textit{MFC}): 
given two depth maps $D^m, D^n \in \cR^{W\times H}$ at frame $m$, $n$ of the video sequence, we unproject $D^m$ into a point cloud;
using the ground-truth world-to-camera poses $P_m,P_n \in \cR^{3 \times 4}$ for frames $m, n$' camera, we transform the point cloud from frame $m$' camera space to frame $n$' camera space,
and project it onto frame $n$'s image plane to yield $D^{m \to n}$. We measure temporal consistency as the average L1 distance between $D^{m \to n}$ and $D^{n}$. We mask out invalid pixels in both frames. 
In practice, we calculate multi-frame consistency on adjacent frames. 

\begin{table*}[t]
\centering
\scalebox{1.1}
{
\setlength{\tabcolsep}{14pt}
\begin{tabular}{lccccccc}
\toprule 
&& Marigold & DAv2 & NVDS & DC & Ours  \\
\cline{3-7}
\multicolumn{2}{c}{Inference Speed (s)} & 5.64 & 0.80 & 1.05 & 1.30 & 0.49 \\
\multicolumn{2}{c}{Compute (GB)}& 5.67 & 23.7 & 20.5 & 8.04 & 6.6  \\
\multicolumn{2}{c}{\# of Parameters (B)} & 1.29 & 0.33 & 0.35 & 2.25 & 2.25  \\
\multirow{2}{*}{Training data}& \# of frames & 74K & 62.6M & 1.4M & - & 39K \\
& \# of scenes & - & - & 14.2K & 203K & 938 \\
\bottomrule
\end{tabular}
}
\vspace{-10pt}
\caption{\textbf{Speed and compute comparison.}}
\label{tab:speed}
\end{table*}

\begin{figure*}[t!]
  \centering
  \includegraphics[clip,trim=0cm 13.6cm 0cm 0cm,width=0.98\linewidth]{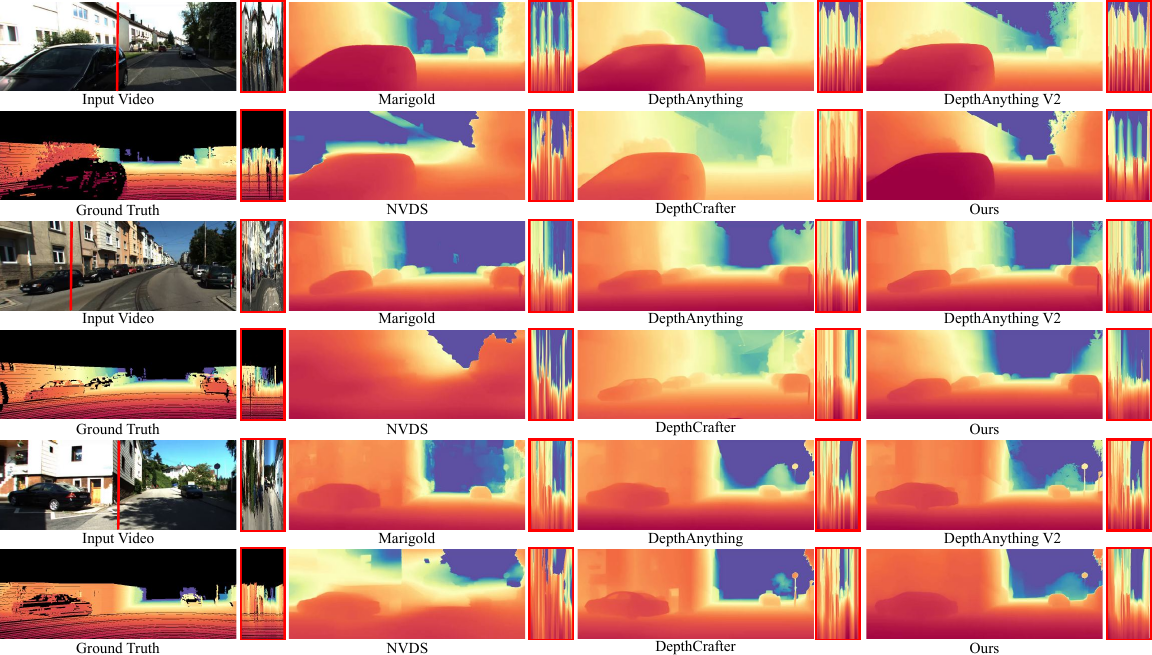}
  \vspace{-0.3cm}
      \caption{\textbf{More qualitative comparison for video depth estimation.} Results on \textit{KITTI-360} datasets.}
    \label{fig:more_qualitative_2}
\vspace{-0.4cm}
\end{figure*}

\begin{figure*}[t!]
  \centering
  \includegraphics[clip,trim=0cm 0cm 0cm 11.5cm,width=0.98\linewidth]{figures/more_qualitative_2.pdf}
  \vspace{-0.3cm}
      \caption{\textbf{More qualitative comparison for video depth estimation.} Results on \textit{ScanNet++} and \textit{Bonn} datasets.}
    \label{fig:more_qualitative_3}
\vspace{-0.4cm}
\end{figure*}

\section{Detailed Proof on Mathematical Rigor and Fluctuating Guidance}
\label{sec:supp_math}
To ensure enough context information, we aim to sample depth latents $\hat{\bz}^{(\bd_{W:F})}$ conditioned on $\hat{\bz}^{(\bd_{0:W})}$, which is $p_\theta(\hat{\bz}^{(\bd_{W:F})} | \hat{\bz}^{(\bd_{0:W})})$.
\textbf{For the replacement trick}, the sampling of $\hat{\bz}^{(\bd_{W:F})}$ follows standard unconditional sampling from $p_\theta(\hat{\bz}^{(\bd_{0:F})}_{t-1} | \hat{\bz}^{(\bd_{0:F})}_t)$, where $\hat{\bz}^{(\bd_{0:F})}_{t} = \left[ \hat{\bz}^{(\bd_{0:W})}_{t}, \hat{\bz}^{(\bd_{W:F})}_{t} \right]$. Crucially, samples $\hat{\bz}^{(\bd_{0:W})}_{t}$ are replaced at each step by exact forward process samples $q(\hat{\bz}^{(\bd_{0:W})}_{t}|\hat{\bz}^{(\bd_{0:W})})$. 
This causes to update $\hat{\bz}^{(\bd_{W:F})}_{t-1}$ using $\hat{\bz}^{(\bd_{W:F})}_{t-1, \theta}(\hat{\bz}^{(\bd_{W:F})}_{t}) \approx \nE_q[\hat{\bz}^{(\bd_{W:F})}_{t-1} | \hat{\bz}^{(\bd_{W:F})}_{t}, \hat{\bz}^{(\bd_{0:W})}_{t}]$, while what is needed instead is $\nE_q[\hat{\bz}^{(\bd_{W:F})}_{t-1} | \hat{\bz}^{(\bd_{W:F})}_{t}, \hat{\bz}^{(\bd_{0:W})}] = \nE_q[\hat{\bz}^{(\bd_{W:F})}_{t-1} | \hat{\bz}^{(\bd_{0:F})}_{t}, \hat{\bz}^{(\bd_{0:W})}_t]+(\sigma_t^2/\alpha_t)\nabla
_{\hat{\bz}^{(\bd_{W:F})}_{t-1}}
log\ q(\hat{\bz}^{(\bd_{0:W})}|\hat{\bz}^{(\bd_{0:W})}_{t})$. The missing second term introduces dynamic guidance variations across sampling steps.
\textbf{As for our context-aware strategy}, we can do conditional sampling from $p_\theta(\hat{\bz}^{(\bd_{0:F})}_{t-1} | \hat{\bz}^{(\bd_{0:F})}_t)$, with $\hat{\bz}^{(\bd_{0:F})}_{t} = \left[ \hat{\bz}^{(\bd_{0:W})}, \hat{\bz}^{(\bd_{W:F})}_{t} \right]$ without forward process $q(\cdot|\cdot)$. As a result, $\hat{\bz}^{(\bd_{W:F})}_{t-1}$ is updated in the direction provided by $\nE[\hat{\bz}^{(\bd_{W:F})}_{t-1} | \hat{\bz}^{(\bd_{W:F})}_{t}, \hat{\bz}^{(\bd_{0:W})}]$.

\section{Speed and Compute Comparison}
\cref{tab:speed} shows runtime, compute and model parameters. \method{} is significantly faster than Marigold~\cite{Ke2023repurposing} and DepthCrafter (DC)~\cite{hu2024depthcrafter}, and requires a fraction of the memory used by DepthAnything v2(DAv2)~\cite{yang2024depth}, thanks to our more lightweight UNet architecture compared with the baselines.

\section{Additional Ablation}
\label{sec:add_ablation}
We investigate the significance of the small noise level $\sigma_\epsilon$ in the context of overlapping frames within arbitrarily long videos. As illustrated in~\cref{fig:ablation_sigma}, an excessively small $\sigma_\epsilon$ results in degraded spatial and temporal performance due to compounded errors. Conversely, an overly large $\sigma_\epsilon$ also leads to diminished spatial and temporal performance. Consequently, we opt for $\sigma_\epsilon=-4$.

\section{More Qualitative Results}
We provide more qualitative comparisons from \textit{KITTI-360}, \textit{ScanNet++} and \textit{Bonn} datasets in~\cref{fig:more_qualitative,fig:more_qualitative_2,fig:more_qualitative_3}. 
First, we highlight the remarkable spatial accuracy achieved by our method, being comparable to or even better than the one by state-of-the-art models. Furthermore, we can notice how the y-t slice by most methods shows high-frequency artifacts, whereas ours is consistently smoother, confirming the superior temporal consistency we achieve. 

\section{Limitation}
Our method is robust to rapid ego-camera motion (\textit{Scannet++}) and long video (KITTI-360). However, we observe a slight degradation in \textit{AbsRel} 
when handling scenes with abundant dynamic objects (\textit{Sintel}). 
We attribute this to the limited moving objects in the training data, which can be extended in the future.

\end{document}